\documentclass[letterpaper, 10 pt, conference]{ieeeconf}  
\IEEEoverridecommandlockouts                              
\usepackage{amsmath}
\usepackage{algorithmic} 
\usepackage{array}
\usepackage{stfloats}
\usepackage{url}
\usepackage{cite}
\usepackage{graphicx}
\usepackage{mathtools}
\usepackage{threeparttable}
\usepackage{amssymb}
\usepackage{bm}
\usepackage{float}
\usepackage{subfigure}
\usepackage{gensymb}
\usepackage{textcomp}
\usepackage{wasysym}
\usepackage[table,dvipsnames]{xcolor}
\usepackage{hyperref}

\usepackage{booktabs}
\usepackage{blkarray,bigstrut}
\usepackage{multirow}
\usepackage{colortbl,hhline}
\usepackage{hyphsubst}
\usepackage{makecell}

\usepackage{setspace}

\renewcommand{\baselinestretch}{0.98}

\title{\LARGE \bf
Three-dimensional Morphological Reconstruction of Millimeter-Scale Soft Continuum Robots based on Dual-Stereo-Vision
}

\author{Tian-Ao Ren, Wenyan Liu, Tao Zhang, Lei Zhao, Hongliang Ren$^{*}$ and Jiewen Lai$^{*}$
\thanks{This work is supported in part by the IdeaBooster Fund Award under grant 3230391; in part by the Faculty Direct Fund under grant 4055213; in part by the MSc Admission Scholarship of the Department of Electronic Engineering, CUHK; in part by the Hong Kong Research Grants Council (RGC) Collaborative Research Fund under grant CRF C4026-21GF; and in part by the Hong Kong Research Grants Council (RGC) Research Impact Fund under grant R4020-22.}
\thanks{T.-A. Ren is with the Department of Mechanical Engineering, Stanford University, CA, USA. He was with the CUHK Shenzhen Research Institute, Shenzhen, China. \tt\small{tianao@stanford.edu}}
\thanks{W. Liu, T. Zhang, H. Ren, and J. Lai are with the Department of Electronic Engineering, The Chinese University of Hong Kong, Shatin, NT, Hong Kong, China.}
\thanks{L. Zhao is with the College of Computer Science and Electronic Engineering, Hunan University, Changsha, China. She is also with the CUHK Shenzhen Research Institute, Shenzhen, China.}
        \thanks{*Corresponding authors. {\tt\small \{hlren,jwlai\}@ee.cuhk.edu.hk}}
}

\begin{document}

\maketitle

\begin{abstract}

Continuum robots can be miniaturized to just a few millimeters in diameter. Among these, notched tubular continuum robots (NTCR) show great potential in many delicate applications. Existing works in robotic modeling focus on kinematics and dynamics but still face challenges in reproducing the robot's morphology---a significant factor that can expand the research landscape of continuum robots, especially for those with asymmetric continuum structures. This paper proposes a dual stereo vision-based method for the three-dimensional morphological reconstruction of millimeter-scale NTCRs. The method employs two oppositely located stationary binocular cameras to capture the point cloud of the NTCR, then utilizes predefined geometry as a reference for the KD tree method to relocate the capture point clouds, resulting in a morphologically correct NTCR despite the low-quality raw point cloud collection. The method has been proved feasible for an NTCR with a 3.5 mm diameter, capturing 14 out of 16 notch features, with the measurements generally centered around the standard of 1.5 mm, demonstrating the capability of revealing morphological details. Our proposed method paves the way for 3D morphological reconstruction of millimeter-scale soft robots for further self-modeling study.

\end{abstract}

\section{INTRODUCTION}

Soft robots, due to their inherent flexibility and compliance, necessitate innovative sensing methodologies distinct from those used in rigid robots. Vision-based approaches have become increasingly prevalent for measuring various characteristics of soft robots, as traditional sensing methods like joint motor encoders and optical markers are not entirely applicable \cite{lee2017soft,hofer2021vision,faris2023proprioception,rong2024vision}. These conventional methods require direct physical interaction with the highly deformable robotic structure, potentially altering its motion and accuracy \cite{hegde2023sensing}.

Several vision-based methods have been developed to overcome these challenges. For example, Gu \textit{et al.} \cite{rong2024vision} introduced a neural network-based shape estimation approach that achieves a high precision with an error margin of 2.91 mm given a soft parallel robot with three 175 mm length actuators. This method demonstrates robustness under uncertain observation conditions, making it a significant advancement for real-time shape estimation of soft parallel robots. Zheng \textit{et al.} \cite{zheng2024vision} proposed a convolutional neural network (CNN) regression approach for online markerless key point estimation of deformable robots. Their method outperforms state-of-the-art techniques by up to 4.5\% in estimation accuracy, proving effective on various deformable structures such as soft robotic arms and soft robotic fish. Additionally, Lee \textit{et al.} \cite{lee2022vision} explored vision-based key point estimation for deformable wires in wire harness assembly automation, effectively addressing challenges posed by ultra-thin objects. In \cite{lai2021verticalized}, the depth vision-based key points estimation is used as the closed-loop positional feedback to enhance the continuum robot's distal tip trajectory tracking, showing the reliability of using the vision-based system for the soft robot's control.
\begin{figure}
    \centering
    \includegraphics[width=0.98\linewidth]{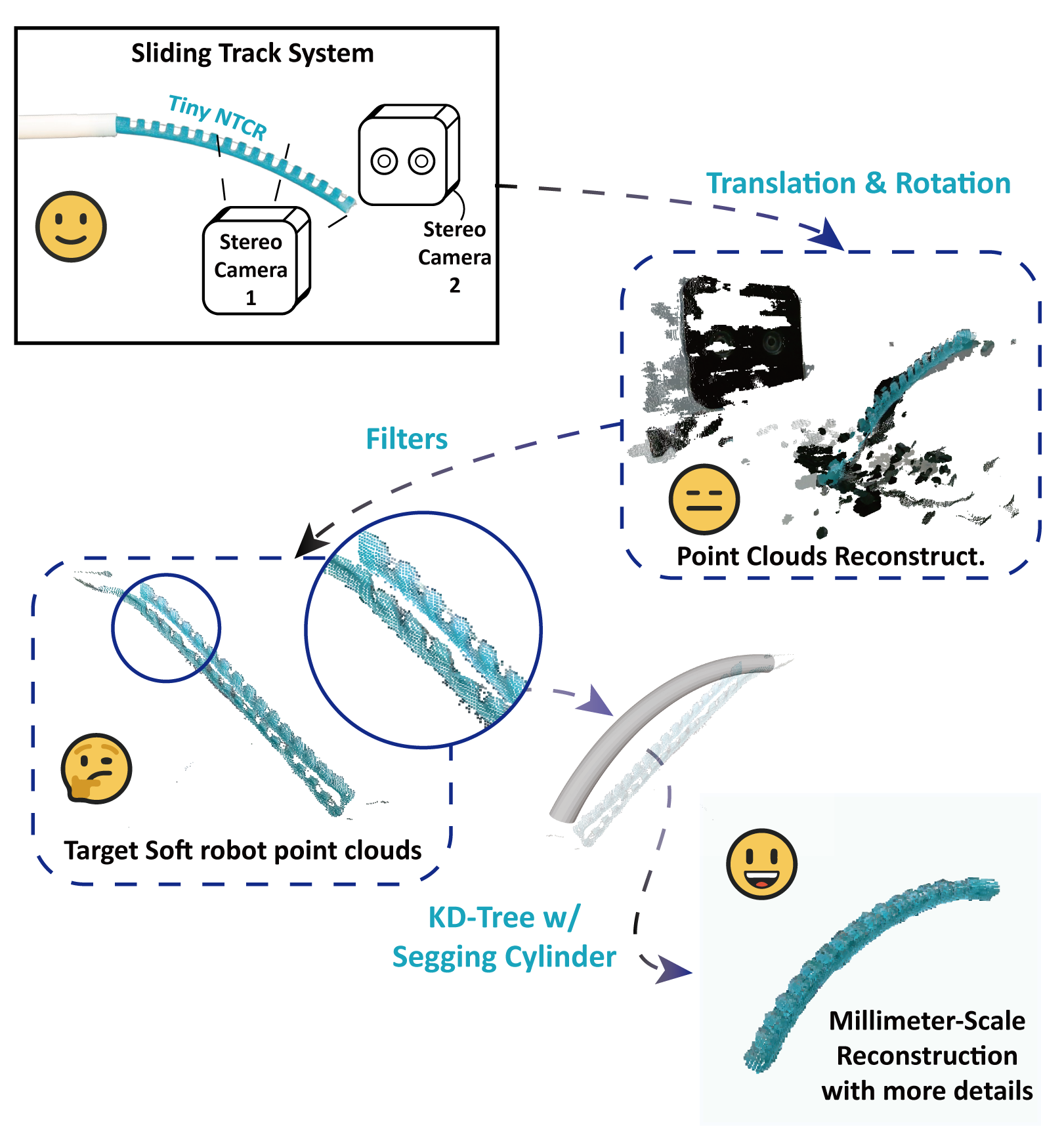}
    \caption{Overview of the 3D-morphological reconstruction and optimization pipeline for millimeter-scale soft robots. It can be seen that the preliminary reconstructed robot demonstrates two flat faces of point clouds due to the low resolution of stereo camera systems. After applying our method, a refined soft robot with more details and patterns can be obtained.}
    \label{fig:overview}
\end{figure}

Despite these advancements, the primary focus of current research has been on key point estimation \cite{centurelli2022closed}, 2D/3D shape \cite{lai2020toward,xu2021visual,dalvand2016high,cui2022coupled} or even tactile \cite{wu2024vision} and force estimation \cite{zhang2018vision} rather than comprehensive morphological reconstruction. While these methods are effective for specific tasks, they do not fully capture the detailed morphology of soft robots. This limitation presents a significant gap in the current research landscape, as a detailed morphological understanding is crucial for many applications, particularly in robot's self-modeling \cite{chen2022fully}.

Current related work has been focusing on shape sensing on soft robots. For instance, Zheng \textit{et al.} \cite{zheng2024vision} proposed a CNN-based key point estimation method that excels in accuracy, but is not designed for detailed morphological reconstruction. Lee \textit{et al.} \cite{lee2022vision} addressed challenges with deformable wires but did not extend their work to the comprehensive morphological reconstruction required for complex soft robots. Our method, by contrast, aims to fill these gaps by providing a practical approach to the full 3D reconstruction of soft robots, addressing the unique challenges associated with their small-scale and complex geometries. The work by Chen \textit{et al.} highlights the benefits of robots creating accurate 3D representations of their morphology and kinematics \cite{chen2022fully}. Similarly, in \cite{zhang2013design}, the tendon-sheath-driven system can enhance search and rescue operations by enabling precise shape reconstruction. These studies collectively advance robotic autonomy and functionality through innovative 3D reconstruction techniques. Moreover, Sim2Real transfer learning for soft robots can benefit from comprehensive high-resolution simulated data to minimize the reality gap—the discrepancy between simulated models and real-world scenarios \cite{lai2023sim}, and employing generated 3D point clouds that could reflect the real-world geometry can potentially enhance the performance and applicability of the existing Sim2Real framework. However, such tools are inaccessible at the current stage.

To address this gap, we propose a novel dual stereo vision-based 3D morphological reconstruction method, particularly for millimeter-scale notched tubular continuum robots (NTCR). Our approach employs two oppositely located stationary binocular cameras to capture the 3D point cloud of the NTCR at a very close distance. Due to the limited camera resolution and small size of the NTCR, the capture point clouds are not organized enough to reveal the true morphology of our target robot. By utilizing predefined geometry as a reference and applying the KD tree method to relocate the captured raw point clouds, these scattered point clouds can be more organized to reveal the robot's morphology as well as the pattern. Figure \ref{fig:overview} provides an overview of the proposed method.

Specifically, this work contributes a novel dual stereo vision-based 3D morphological reconstruction method for small notched tubular continuum robots in point clouds form. By leveraging the predefined geometry, we propose a point clouds relocating strategy that involves filtering and KD tree algorithms, streamlining a point cloud-based morphology reconstruction protocol for small continuum robots with a diameter below 3.5 mm. Our method has been rigorously verified with experiments, showcasing the performance in detail morphological reconstruction in comparison to the raw point clouds. Our method provides an alternative to not using expensive point cloud-capturing systems for small objects, which could be used to model other small robots or objects with not just shapes but delicate details.






\section{Methodology}

\subsection{Capturing Raw 3D Point Clouds}
\begin{figure}[t!]
\centering
\includegraphics[width=0.6\linewidth]{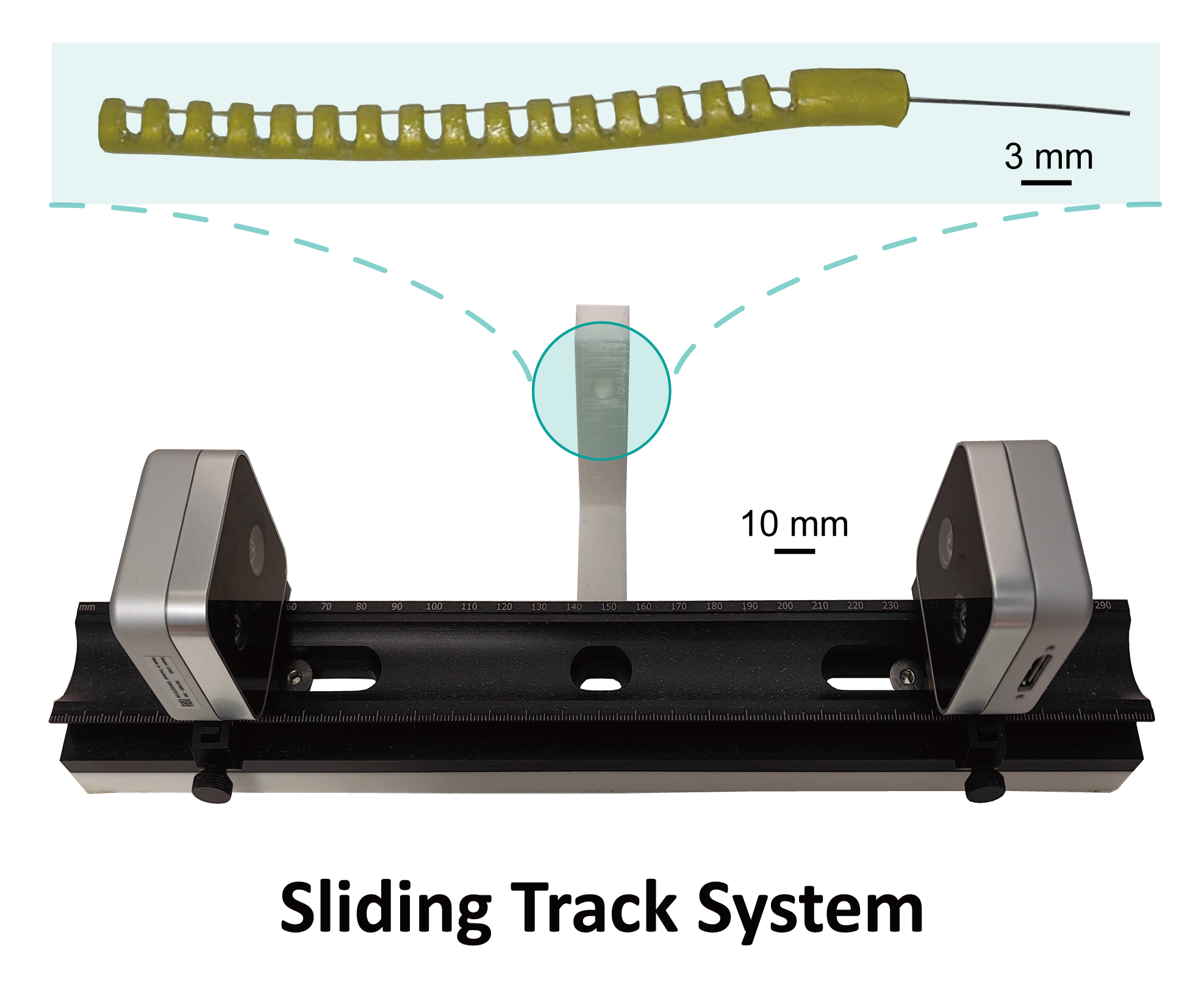}
\caption{The experimental setup with the sliding track system and dual Intel RealSense D405 cameras.}
\label{fig:experimental_setup}
\end{figure}
To capture the raw 3D point clouds for the NTCR, as shown in Fig. \ref{fig:experimental_setup}, we built a dual-stereo-vision system by utilizing two Intel RealSense D405 cameras. These two cameras are arranged at a calibrated baseline distance to capture the opposite perspectives of the object (i.e., the small continuum robot). Although the system can become more explicit by using multiple cameras \cite{chen2022fully}, we opt for the minimal required visions to reduce the data redundancy. These cameras generate depth maps, which provide critical spatial information about the object's shape and rough appearance. The depth \( Z \) of a point \( P \) can be derived from the disparity \( d \) between the corresponding points in the left and right camera images, i.e.,
\begin{equation}
d = |X_L - X_R|,
\end{equation}
\begin{equation}
Z = \frac{bf}{d},
\end{equation}
where, \( b \) is the baseline distance between the two cameras; \( f \) is the focal length of the cameras; and \( d \) is the disparity.

After that, the depth maps undergo the processing steps, including intrinsic and extrinsic camera calibration to correct distortions and accurately map the 3D coordinates. Subsequently, point cloud data is extracted from the depth maps, forming the basis for further refinement. This structured approach ensures precise 3D reconstruction, which is essential for applications requiring high-fidelity models \cite{zhang2000flexible}.

\subsection{Conversion and Filtering}
\label{filtering}
The conversion and filtering stage is pivotal for refining raw data obtained from 3D scanning to produce accurate 3D models. Depth maps captured by the dual-stereo-vision system are first converted into point cloud data \cite{rusu20113d}. The transformation from pixel coordinates \((u, v)\) to camera coordinates \((X_c, Y_c, Z_c)\) is given by the intrinsic camera matrix \( K \):
\begin{equation}
\begin{bmatrix}
u \\
v \\
1 
\end{bmatrix}
= 
K
\begin{bmatrix}
X_c / Z_c \\
Y_c / Z_c \\
1 
\end{bmatrix},
\end{equation}
where \( K \) is the intrinsic camera matrix defined by
\begin{equation}
K = 
\begin{bmatrix}
f_x & 0 & c_x \\
0 & f_y & c_y \\
0 & 0 & 1 
\end{bmatrix}.
\end{equation}
This conversion involves transforming 2D pixel coordinates into 3D spatial coordinates using parameters from the calibrated cameras. Once the point clouds are generated, they undergo several filtering processes to enhance data quality. Here, Statistical Outlier Removal (SOR) \cite{bustos2017guaranteed} is utilized to identify and exclude points that significantly deviate from their neighbors, thus preserving data integrity. For each point, the mean distance \( \mu \) to all its neighbors is computed. Points are considered outliers if their mean distance exceeds a threshold \( \mu + \alpha \sigma \), where \( \sigma \) is the standard deviation of distances:
\begin{equation}
\delta_{\text{Threshold}} = \mu + \alpha \sigma.
\end{equation}
On top of that, three conditional filters are applied to retain points meeting specific criteria, such as (1) spatial coordinates and (2) color values. To further smooth the data, the (3) Moving Least Squares (MLS) filter is employed, fitting local polynomial surfaces to the point cloud to reduce noise while maintaining geometric details. Notably, the MLS filter smooths the point cloud by fitting a polynomial to the points within a neighborhood. The fitted function $J$ minimizes the weighted sum of squared distances from the points to the polynomial surface, viz.,
\begin{equation}
J = \sum_{p \in \mathbb{N}_{>0}} w \left( \| p - x \|^2 \right) \| f(x) - p \|^2,
\end{equation}
where $w \left( \| p - x \|^2 \right)$ is a weight function that decreases with distance.

\subsection{Triangular Patch Construction}
Triangular patch construction is essential for converting filtered point cloud data into a continuous and coherent 3D mesh \cite{curless1996volumetric}. This process involves identifying neighboring points within the point cloud and connecting them to form triangles, thereby creating a mesh that represents the object’s surface. Poisson surface reconstruction \cite{kazhdan2006poisson} is a prevalent technique used in this step, which involves solving the Poisson equation to balance surface smoothness with fidelity to the original point data. The key equations is given by
\begin{equation}
\Delta \chi = \nabla \cdot \mathbf{V},
\end{equation}
where \( \Delta \) is the Laplace operator, measuring how a function diverges from its average value; \( \chi \) is the implicit function representing the surface; \( \nabla \cdot \) denotes the divergence operator; and \( \mathbf{V} \) is the vector field derived from the point cloud data. And the energy function \( E(\chi) \) to be minimized is:
\begin{equation}
E(\chi) = \int \|\nabla \chi - \mathbf{V}\|^2 \, \mathrm{d}p,
\end{equation}
where, \( E(\chi) \)is the energy function representing the total error between the gradient of \( \chi \) and \( \mathbf{V} \); \( \|\nabla \chi - \mathbf{V}\|^2 \) is the squared norm of the difference, representing the local error at each point; and \( \mathrm{d}p \) is the differential element of the domain.

This method generates a `watertight' surface by interpreting the point cloud as samples of an implicit function, effectively filling gaps and ensuring continuity. Poisson reconstruction is particularly adept at handling noisy and incomplete data, as it robustly reconstructs the underlying surface by considering the normal vectors of the points. The resulting triangular mesh accurately depicts the robot’s geometry, providing a high-quality 3D model suitable for visualization, and detailed analysis.

\section{Experiment}

To achieve accurate 3D reconstruction of millimeter-scale NTCR, a comprehensive experimental setup was devised. This section outlines the detailed experimental setup, the comparison of depth cameras used, and the preliminary results obtained from the scanning and reconstruction process.

\subsection{Experiment Setup}
\label{setup_exp}
The experimental setup for 3D scanning and reconstruction of millimeter-scale NTCR involves a custom-designed sliding track system equipped with two Intel RealSense D405 depth camerasThe cameras are mounted on adjustable slides that can move along the track, facilitating the calibration and optimization of their relative positions. A 3D-printed rack is placed in between the cameras that allows the NTCR to maintain a fixed position at the robot base, and this base frame can be referred to as the reference frame. The base features a vertical post with small holes for anchoring the robot, ensuring its stability and preventing movement during scanning. This arrangement is crucial for maintaining consistent reference coordinates throughout the scanning process, particularly when the robot’s posture is altered by manipulating an internal control wire. As such, a point cloud data collection platform is established. The setup is shown in Figure \ref{fig:experimental_setup}. This setup ensures that the reference coordinates of the robot remain consistent, even when its configuration is altered by manipulating an internal control wire.

\subsection{Depth Cameras Comparison and Calibration}

To determine the optimal depth camera for accurate 3D reconstruction, both the Intel RealSense D435i and D405 cameras were evaluated. The cameras were assessed based on their ability to capture detailed point clouds of the same NTCR within a diameter of 3.5 mm (with 1.5-mm width notches), focusing on measurables such as noise levels, resolution, and the clarity of structural features.

\begin{figure}
    \centering
    \includegraphics[width=0.92\linewidth]{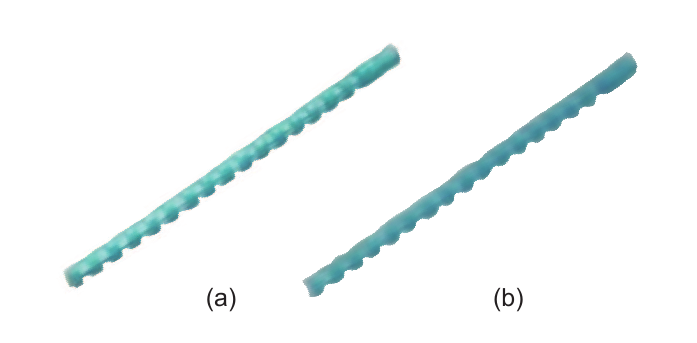}
    \caption{Comparison of Point clouds (with RGB information) captured by Intel RealSense (a) D405 and (b) D435i.}
    \label{fig:comparison_435i_405}
\end{figure}

The D435i camera exhibited certain limitations, producing point clouds with significant noise and incomplete details, failing to clearly distinguish the robot from the surrounding environment. Conversely, the D405 camera demonstrated better performance, generating point clouds with clear edges, well-defined boundaries, and high-resolution details, particularly capturing the grooves and intricate structures of the robot. The comparison can be seen in Fig. \ref{fig:comparison_435i_405}. However, according to the technical specification, the depth accuracy of D405 is below 2\% at a range of 50 cm, meaning that it could result in an error of 10 mm. Although its performance has been the best among other depth cameras like Helios 2, Azure Kinect, and RealSense D435i \cite{hong2024real}, a 10 mm error in depth estimation would produce poor-quality point clouds for miniature objects. Based on this comparative analysis, the D405 camera was selected for our point cloud data collection platform due to its ability to produce the best depth data with minimal noise.

The extrinsic parameters, which include the rotation matrix \( R \) and the translation vector \( t \), transform coordinates from the world coordinate system to the camera coordinate system:
\begin{equation}
\begin{bmatrix}
X_c \\
Y_c \\
Z_c 
\end{bmatrix}
= 
R 
\begin{bmatrix}
X_w \\
Y_w \\
Z_w 
\end{bmatrix}
+ t,
\label{eq_tr}
\end{equation}
where $R$ and $t$ can be easily obtained thanks to the adjustable experimental setup as introduced in Sec. \ref{setup_exp}.

\subsection{Preliminary Results}

Preliminary scanning was performed using our setup with two cameras positioned at an optimal distance of 90 mm from the target continuum robot. Depth maps were simultaneously captured from two perspectives, and point cloud models were generated from each camera’s depth data. These point clouds were then aligned and merged using a motion matrix that incorporated the cameras’ relative positions and orientations using Eq. \eqref{eq_tr}. Initial data filtering involved removing noise and outliers from the point clouds using Statistical Outlier Removal (SOR) \cite{rusu2013semantic, rusu20113d} and conditional filters as introduced in Sec. \ref{filtering}. These filtering techniques significantly improved the data quality. Subsequently, Poisson surface reconstruction was applied to the filtered point clouds to create a continuous and coherent 3D mesh.

\begin{figure}[t!]
\centering
\includegraphics[width=1\linewidth]{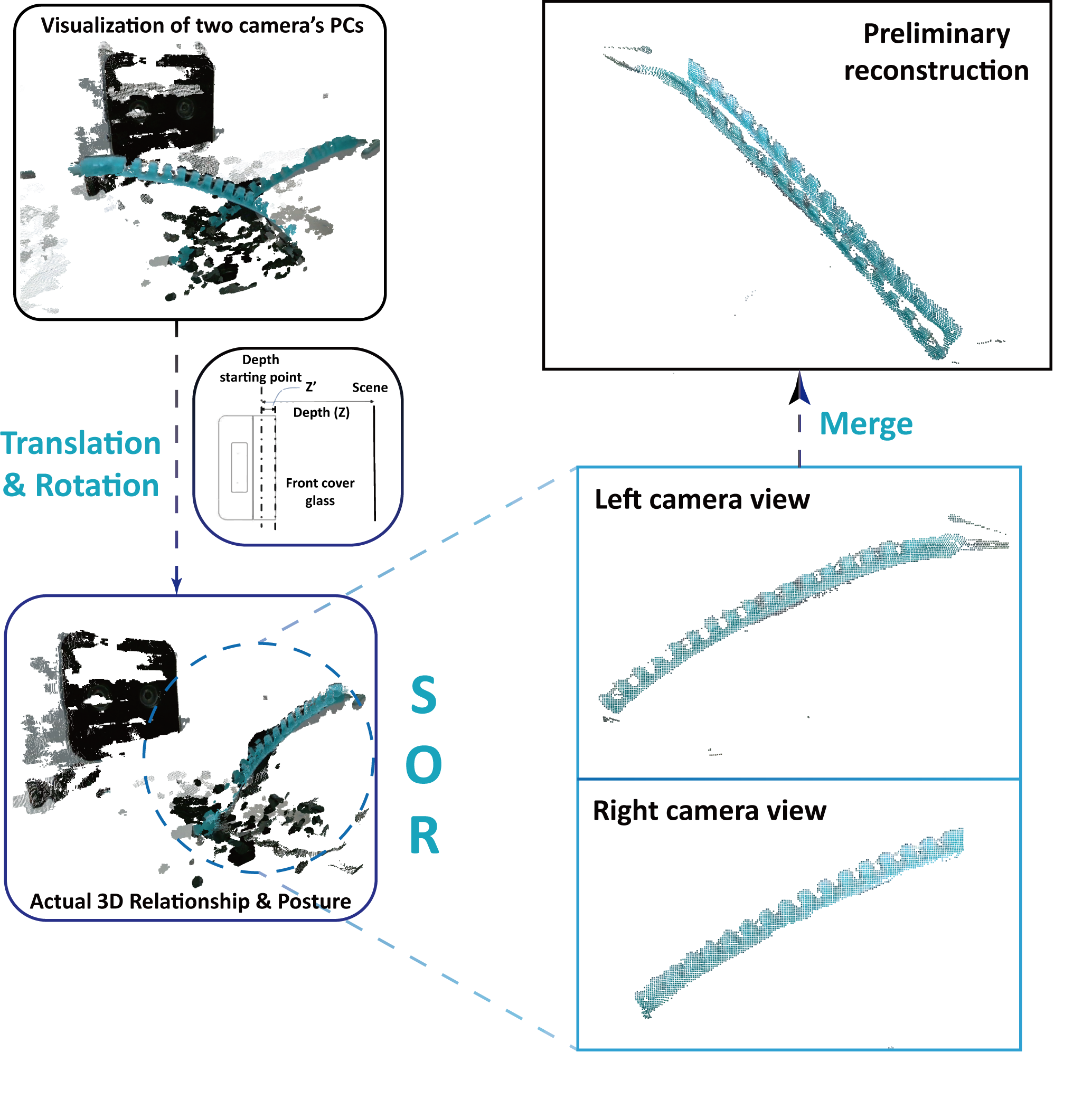}
\caption{Preliminary 3D reconstruction results showing the detailed geometry of the NTCR.}
\label{fig:preliminary_results}
\end{figure}

The preliminary results, shown in Figure \ref{fig:preliminary_results}, demonstrated that our point cloud data collection platform and processing pipeline could effectively capture and reconstruct the detailed geometry of the millimeter-scale soft robot. The resultant 3D model accurately represented the robot’s basic structural features, including its curvature and fine notches/grooves. However, it can also be noticed that the preliminary result shows two separated point cloud sets as they were captured by two cameras in the opposite direction. The cameras' insufficient accuracy in the depth direction also contributes to such a result. The preliminary reconstructed point clouds fail to represent the cylindrical appearance of the tiny robot, which awaits further optimization.


\section{Optimization}

Additional optimization techniques were employed to further refine the 3D reconstruction of the NTCR. This section details the use of KD-Tree for efficient point cloud processing and presents the results with analysis and discussion.

\subsection{KD-Tree}

The KD-Tree (K-Dimensional Tree) \cite{bentley1975multidimensional} is a spatial data structure that facilitates efficient nearest neighbor searches, which is useful for processing and optimizing large point cloud datasets. In this study, the KD-Tree was utilized to refine the alignment and accuracy of the reconstructed preliminary NTCR's 3D models. This model can be either known in prior, or estimated from the sparse point clouds through special regression or curve fitting. In this method, the Euclidean distance \( d(p, q) \) between two points \( p \) and \( q \) in 3D space is defined as:
\begin{equation}
d(p, q) = \sqrt{(p_x - q_x)^2 + (p_y - q_y)^2 + (p_z - q_z)^2}.
\end{equation}
To integrate the KD-Tree into our workflow, the following steps were undertaken:
\paragraph{Construction}
The KD-Tree was constructed using the filtered point cloud data obtained from the initial 3D reconstruction. Each point in the point cloud was inserted as a node in the KD-Tree. This structure enabled efficient spatial querying, which is essential for subsequent processing steps.
\paragraph{Nearest Neighbor Search}
For each point representing the current scan of the NTCR, the KD-Tree was employed to find the nearest corresponding point in the reference point cloud representing the ideal geometric model \cite{muja2014scalable}. The nearest neighbor search minimizes the $d(p,q)$ between the matched points.
\paragraph{Point Cloud Alignment}
Using the nearest neighbor pairs identified through the KD-Tree, an iterative closest point (ICP) algorithm \cite{rusinkiewicz2001efficient} was applied to align the target point cloud to the reference point cloud. The ICP algorithm aims to minimize the difference between corresponding points in two point clouds \cite{best1992method}. It iteratively solves for the transformation matrix \( T \) as
\begin{equation}
T = \arg \min_T \sum_i \| p_i - (R q_i + t) \|^2
\end{equation}
where, \( p_i \) denotes the points in the target point cloud; and \( q_i \) denotes the corresponding points in the reference point cloud. This alignment involved transforming the target points to minimize the overall distance to their corresponding reference points, improving the coherence of the 3D model.

\paragraph{Iterative Refinement}
The alignment process was refined through multiple iterations, continuously updating the KD-Tree with the latest transformed point cloud data. This iterative approach ensured that the alignment errors were progressively reduced, resulting in a highly accurate final model.

\begin{figure}[t!]
\centering
\includegraphics[width=1\linewidth]{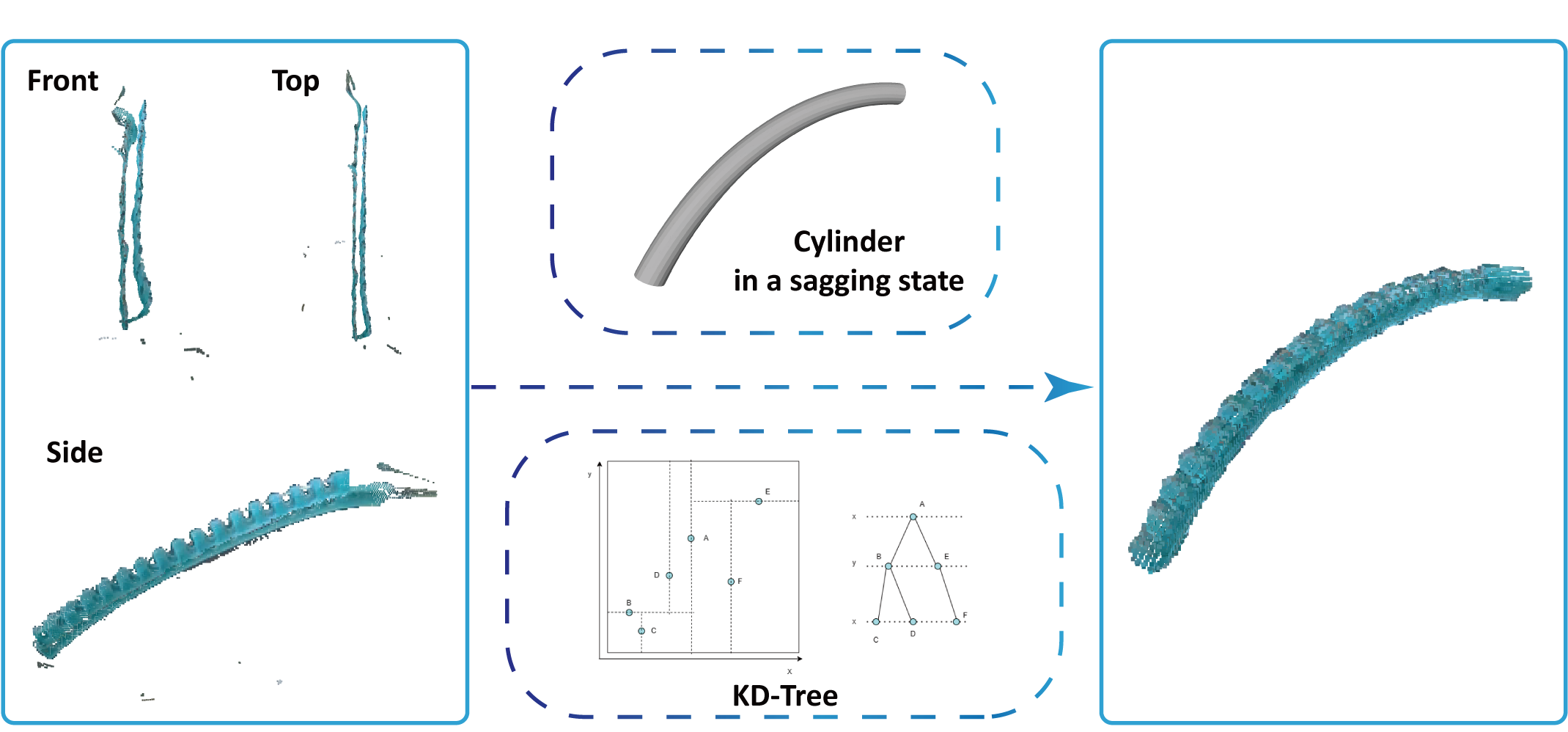}
\caption{Optimized 3D model of the NTCR after KD-Tree-based alignment and refinement. It can be seen that, compared to the preliminary result, the optimized model has no flat surfaces due to the re-alignment of the point clouds. The notch patterns are clearer.}
\vspace{-.4cm}
\label{fig:optimized_3d_model}
\end{figure}

Figure \ref{fig:optimized_3d_model} demonstrates the optimized 3D model of the example NTCR following KD-Tree-based alignment and refinement, showcasing the improved structural details and accuracy.

\subsection{Analysis \& Discussion}
We compare the point density consistency (PDC) \cite{rusu20113d} between two point clouds: one from our final optimized algorithm (Point Cloud 1) and another from the non-optimized algorithm before KD-Tree mapping (Point Cloud 2). The metric PDC measures the uniformity of point distribution in a 3D point cloud. It quantifies how evenly the points are spread throughout the desired grid/voxel space. PDC can be computed by
\begin{equation}
    \lambda_{\textrm{PDC}}= \frac{\sigma_{\textrm{PDC}}}{\mu_{\textrm{PDC}}},
\end{equation}
where $\sigma_{\textrm{PDC}}$ and $\mu_{\textrm{PDC}}$ are the standard deviation of point counts per voxel and the mean of point counts per voxel, respectively. Here, the voxel size for point density calculation is set to 0.5 mm. The mean density and $\sigma_{\textrm{PDC}}$  for the point clouds are summarized in the following Table \ref{point density calculation}.
\begin{table}[t!]
\caption{Point Density Consistency Analysis}
\footnotesize
\begin{center}
\rowcolors{2}{lightgray!30}{white}
\begin{tabular}{cccc}
\hline
 \rowcolor{blue!12}\textbf{Point Clouds} & $\mu_{\textrm{PDC}}$ & $\sigma_{\textrm{PDC}}$ & $\lambda_{\textrm{PDC}}$ \\
\hline
Point Clouds 1 (Optimized)  & 1.5489 & 3.9636 & 2.5590\\
Point Clouds 2 (Non-optimized)  & 0.9582 & 2.2759 & 2.3752\\
\hline
\end{tabular}
\end{center}
\label{point density calculation}
\end{table}
The optimized algorithm generates point clouds with a higher average point density $\mu_{\textrm{PDC}}$, which contributes to a more detailed model. However, the increased standard deviation suggests that there is more variability in point placement $\sigma_{\textrm{PDC}}$. Depending on the application, this variability can either be beneficial (capturing more detailed features) or detrimental (introducing noise). Further refinement of the optimization algorithm could aim to balance high mean density with lower variability, enhancing both the accuracy and consistency of the point cloud.

\begin{figure}[t!]
\centering
\includegraphics[width=1\linewidth]{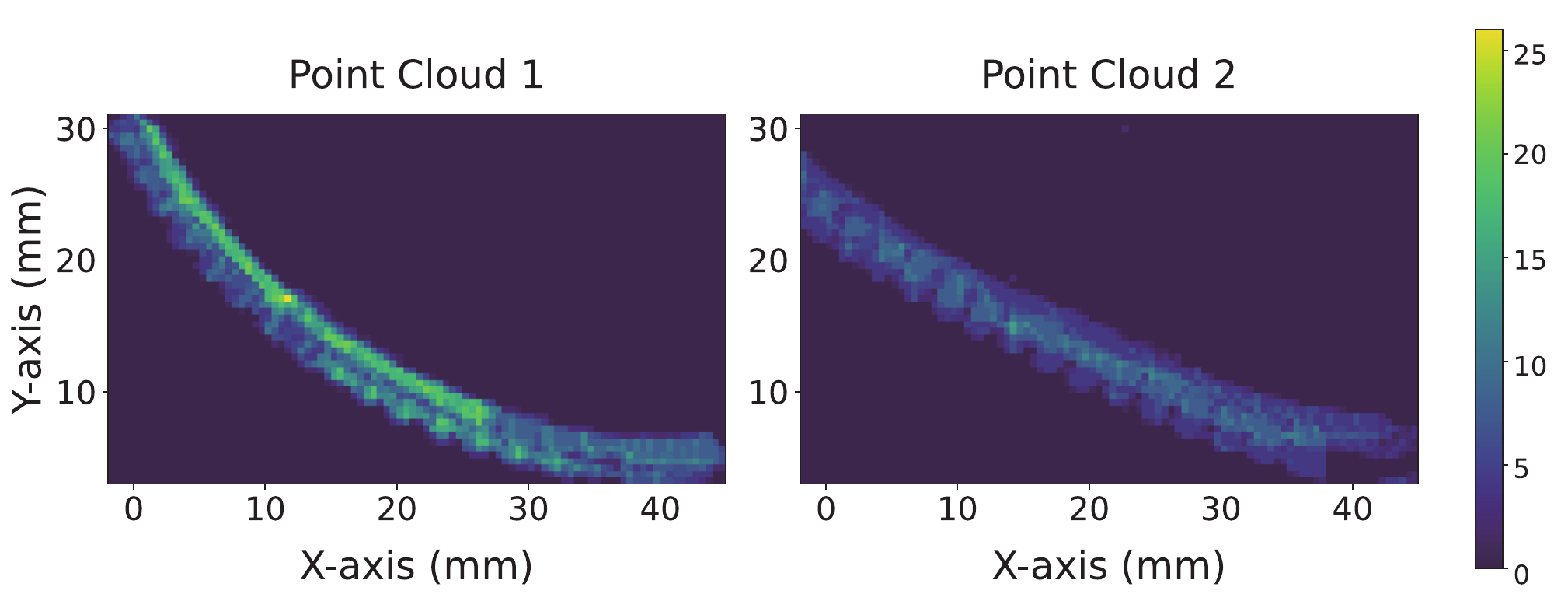}
\caption{Point density heatmaps for Point Cloud 1 (Optimized) and Point Cloud 2 (Non-optimized). The color bar represents point density per grid cell (2D), with higher values (color bar, unit: number of points) indicating denser regions.}
\vspace{-.4cm}
\label{fig:heatmap}
\end{figure}

Our algorithm improves the point cloud density distribution without significantly increasing the computational load $\lambda$, thereby capturing more necessary details and enhancing the overall quality of the 3D model. Our result underscores the need for both high point density and uniform distribution to improve point cloud data quality. The following heatmaps, as shown in Fig. \ref{fig:heatmap} illustrate the point density distribution for both point clouds, with higher density regions in Point Cloud 1 (left) compared to Point Cloud 2 (right). We can see that the optimized model exhibits well-defined edges and an accurate representation of the NTCR's structural features, including notches and curvilinear profile. This was corroborated by quantitative metrics such as surface roughness and curvature consistency, which showed marked improvements.

\begin{figure}[t!]
\centering
\includegraphics[width=1\linewidth]{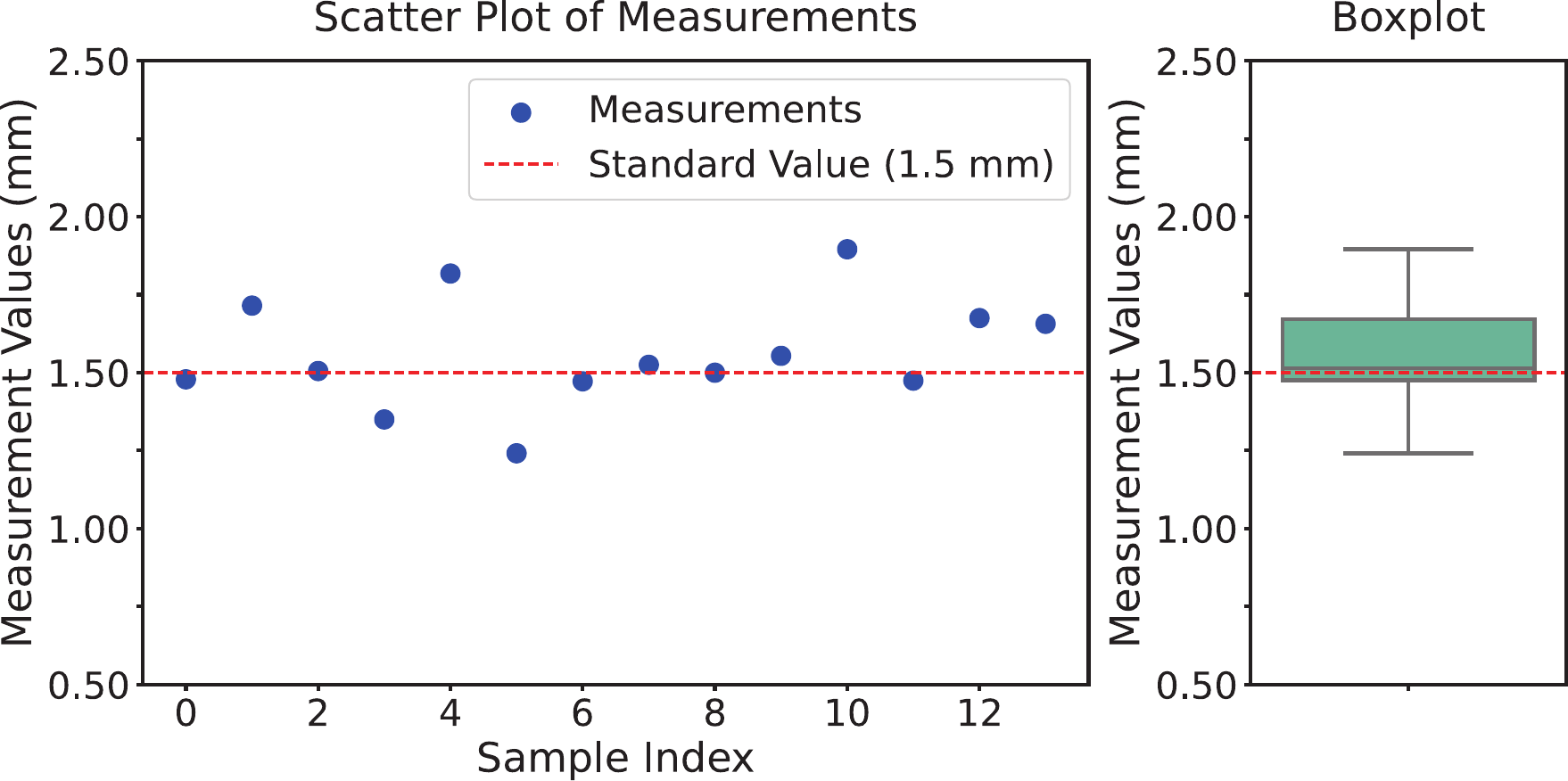}
\caption{Measurement results on the visible notches' width, highlighting the comparison with a standard reference width of 1.5 mm. (left) The scatter plot shows individual measurement variations, while the (right) boxplot provides a summary of the width distribution.}
\vspace{-.3cm}
\label{fig:boxplot}
\end{figure}

As shown in Fig. \ref{fig:boxplot}, the measurements for the notches are generally centered around the standard width of 1.5 mm. The variability is within acceptable limits, with most measurements not deviating significantly from the standard. In our reconstructed structure, out of 16 notches, 14 were observable and measurable in the final results. The result highlights individual variations in the widths of these notches, with some measurements notably above or below the standard value. The boxplot summarizes the data distribution, indicating that most measurements fall within a consistent range around the median.  Additionally, the natural bending of the robot during measurement could influence the notch widths, making the overall results quite satisfactory. The statistical data for the 14 measured notches is shown below, demonstrating the overall accuracy and precision of our reconstruction method.

The detailed analysis confirms that the KD-Tree-based optimization effectively enhances the 3D reconstruction process by improving accuracy and completeness and reducing noise. These improvements result in a high-fidelity model that accurately represents the intricate appearance of millimeter-scale soft robots, thereby advancing the capabilities for precise surgical interventions and other critical applications.

\section{Conclusions}

In this paper, we proposed a dual stereo vision-based method for the three-dimensional morphological reconstruction of millimeter-scale NTCRs. We developed a dual stereo vision-based point clouds data collection platform specifically for small NTCRs, and it might be applicable to other small objects. A series of filtering and optimization techniques were used to refine the sparsely located 3D point clouds to be more organized with the aim of reconstructing the robot's morphological appearance despite the hardware limitation.

Our work addresses the unique challenges posed by the small-scale and complex geometries of these robots, providing a useful foundation for 3D point cloud collection and post-processing techniques. Future work will focus on real-time reconstruction capabilities and further refinement of the models to enhance their applicability in dynamic and interactive environments, as well as vision-based self modeling for soft robots.

\bibliographystyle{IEEEtran}

\bibliography{ref}

\end{document}